\documentclass[a4paper,fleqn]{cas-sc}

\usepackage[authoryear,longnamesfirst]{natbib}

\def\tsc#1{\csdef{#1}{\textsc{\lowercase{#1}}\xspace}}
\tsc{WGM}
\tsc{QE}
\tsc{EP}
\tsc{PMS}
\tsc{BEC}
\tsc{DE}

\begin{document}
\let\WriteBookmarks\relax
\def\floatpagepagefraction{1}
\def\textpagefraction{.001}

\shorttitle{HERMES for Long-Tail Autonomous Driving}

\shortauthors{Tang et al.}

\title[mode=title]{\textbf{HERMES}: A \textbf{H}olistic \textbf{E}nd-to-End \textbf{R}isk-Aware \textbf{M}ultimodal \textbf{E}mbodied \textbf{S}ystem with Vision--Language Models for Long-Tail Autonomous Driving}              



%


\author[1]{Weizhe Tang}

\author[1]{Junwei You}[orcid=0009-0002-6447-8276]

\cormark[1]

\ead{jyou38@wisc.edu}

\author[1]{Jiaxi Liu}

\author[1]{Zhaoyi Wang}

\author[1]{Rui Gan}

\author[1]{Zilin Huang}

\author[2]{Pei Li}

\author[1]{Sikai Chen}

\author[3]{Feng Wei}

\author[1]{Bin Ran}

\begin{abstract}
End-to-end autonomous driving models increasingly benefit from large
vision--language models for semantic understanding, yet safe and reliable
planning under long-tail conditions remains challenging, particularly in
mixed-traffic environments involving heterogeneous road users and rare
safety-critical interactions. This paper proposes HERMES, a holistic
risk-aware end-to-end multimodal driving framework that explicitly
incorporates long-tail semantic knowledge into trajectory planning. HERMES
employs a foundation-model-assisted annotation pipeline to construct
structured \textit{Long-Tail Scene Context} and \textit{Long-Tail Planning
Context}, capturing hazard-centric scene information, maneuver intent, and
risk-aware planning guidance. A \textit{Tri-Modal Driving Module} then
integrates multi-view visual observations, historical ego-motion, and
long-tail semantic instructions through intent- and risk-aware conditioning
for trajectory generation. Extensive experiments on a large-scale real-world
long-tail driving benchmark demonstrate consistent improvements over
representative recent baselines in overall planning performance and across
diverse safety-critical scenarios. Ablation studies further validate the
effectiveness and complementary roles of the major components within HERMES.
\end{abstract}


\affiliation[1]{
    organization={Department of Civil and Environmental Engineering, University of Wisconsin--Madison}, city={Madison}, state={WI}, postcode={53706}, country={USA}}


\affiliation[2]{organization={Department of Civil and Architectural Engineering and Construction Management, University of Wyoming},
    city={Laramie},
    postcode={82071},
    state={Wyoming},
    country={United States}}

\affiliation[3]{
    organization={Chongqing University},
    city={Chongqing},
    postcode={400044},
    country={China}
}


\cortext[cor1]{Corresponding author.}

\begin{highlights}
\item We propose HERMES, a holistic framework that leverages large foundation models to systematically address safety-critical long-tail scenarios in end-to-end autonomous driving, where the foundation model is specifically applied to identify and reason about rare corner cases that are critical for safe trajectory planning and vehicle operation.

\item We construct a specialized dataset for long-tail scenario reasoning based on an existing real-world long-tail end-to-end driving benchmark. On top of the benchmark, we add two types of fixed offline annotations: the \textit{Long-Tail Scene Context} for multi-view hazard-centric descriptions, and the \textit{Long-Tail Planning Context} including the risk level, driving intention, high-level directives, and planning rationale, which provides an explicit foundation for risk-informed planning and safety-critical decision-making in long-tail scenarios.

\item We introduce a multimodal fusion architecture, termed the \textit{Tri-Modal Driving Module}. It delivers risk-aware and accurate end-to-end trajectory planning through the fusion of multi-view camera features, historical ego-motion patterns, and long-tail instruction embeddings encoded from the long-tail contexts.

\item Extensive experiments on the annotated dataset demonstrate the effectiveness of our approach against existing methods, with ablation studies further validating the contribution of each key component.
\end{highlights}

\begin{keywords}
Autonomous driving \sep
Long-tail scenarios \sep
Vision--language models \sep
Trajectory planning \sep
Multimodal learning
\end{keywords}

\maketitle

\section{Introduction}

The rapid advancement of artificial intelligence (AI) has catalyzed transformative progress in autonomous driving, where ensuring safety through accurate perception, prediction, and decision-making remains the paramount challenge (\cite{badue2021self, schwarting2018planning, huang2024human}). In particular, these challenges are exacerbated in long-tail mixed-traffic environments, where autonomous vehicles must interact with human-driven vehicles and vulnerable road users under complex, uncertain, and socially coupled conditions.

Early autonomous driving systems predominantly adopted a modular architecture, decomposing 
the complex driving task into sequential components: perception, prediction, planning, 
and control (\cite{yurtsever2020survey, thrun2006stanley, urmson2008boss,  you2025followgen, gan2025goal, ma2025safety}), where each module could be independently developed and optimized. For instance, perception systems detect objects and lane boundaries, while prediction modules forecast agent trajectories using probabilistic frameworks such as MultiPath (\cite{chai2019multipath}), graph-based representations like VectorNet (\cite{gao2020vectornet}), and goal-driven approaches such as TNT (\cite{zhao2021tnt}). The modular approach enables clear failure attribution and facilitates leveraging domain expertise for targeted improvements. However, this decomposition introduces fundamental limitations: information loss may occur at module boundaries as rich sensor data is compressed into intermediate representations, errors propagate and compound through the pipeline, and hand-crafted interfaces prevent joint optimization toward the ultimate driving objective (\cite{bojarski2016end, chen2015deepdriving}).

\begin{figure}[]
  \centering
  \includegraphics[width=0.7\textwidth]{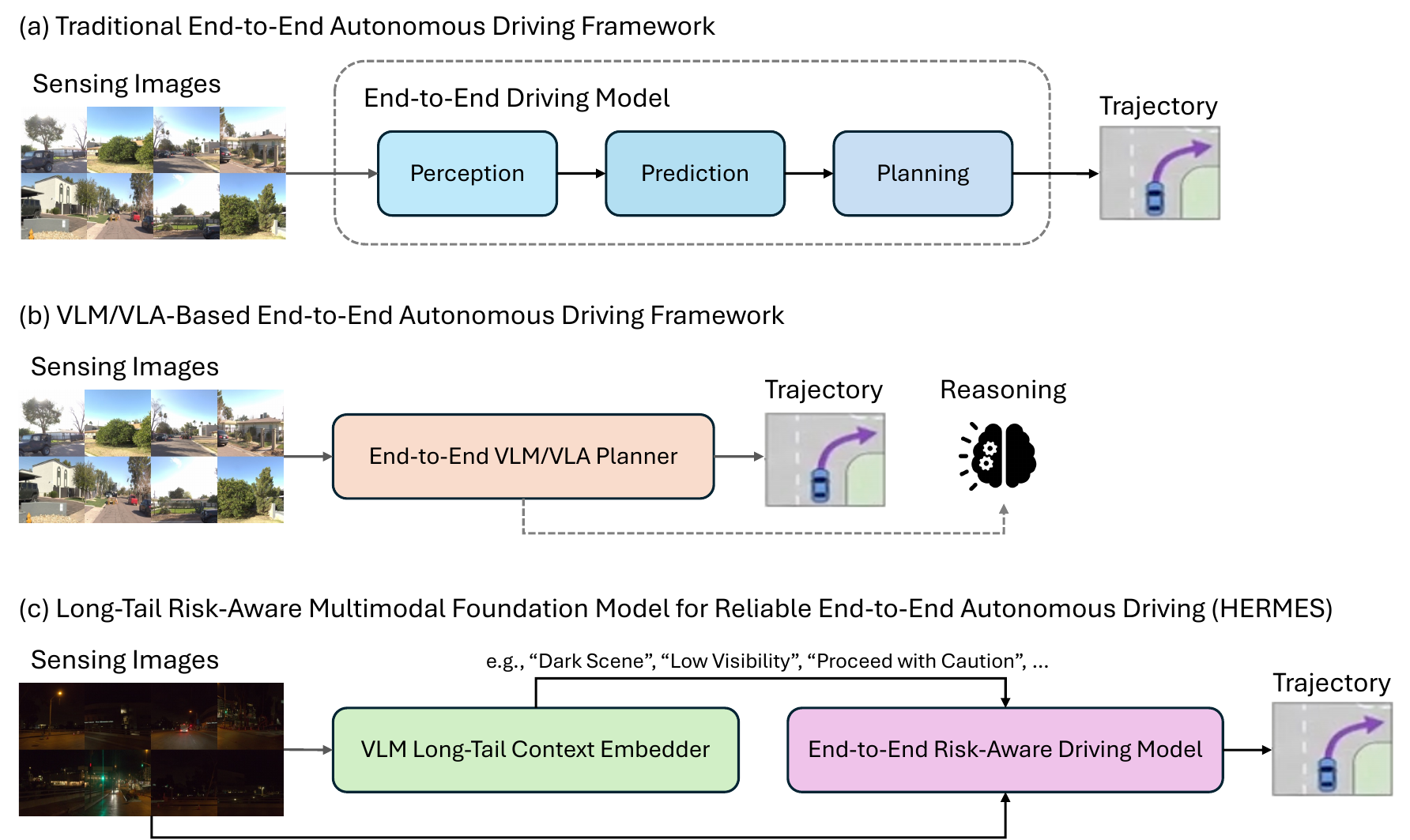}
  \caption{Comparison of autonomous driving paradigms. (a) Traditional end-to-end models 
  integrate perception, prediction, and planning in a unified pipeline. (b) VLM/VLA-based 
  approaches leverage foundation models for reasoning and planning. (c) HERMES (ours) combines 
  VLM-generated long-tail context embeddings with a risk-aware end-to-end driving model for safe trajectory planning.}
  \label{fig:intro}
\end{figure}

To address these information bottlenecks, advances have explored planning-oriented end-to-end frameworks that enable joint optimization across all driving tasks through differentiable training. Specifically, InterFuser (\cite{shao2022interfuser}) fuses multimodal multi-view sensor data through transformer-based attention mechanisms, generating interpretable semantic outputs such as waypoints and object density maps to constrain control predictions. UniAD (\cite{hu2023uniad}) integrates full-stack driving tasks including tracking, mapping, motion forecasting, and occupancy prediction within a unified query-based transformer architecture, enabling gradient flow from planning objectives back through all preceding modules. VAD (\cite{jiang2023vad}) employs a fully vectorized scene representation, encoding dynamic agents and static map elements as explicit instance-level planning constraints while achieving remarkable inference efficiency. These systems achieve state-of-the-art planning 
performance by preserving continuous information flow through differentiable architectures. 
However, a critical limitation emerges. While these models excel at pattern recognition from large-scale training data, they lack the high-level semantic scene understanding and common-sense reasoning capabilities necessary to interpret complex traffic scenarios, understand implicit social conventions, and generalize to rare corner cases beyond their training distribution.

Recognizing this semantic understanding gap, further studies have turned to large language 
models (LLMs) and vision-language models (VLMs) to leverage their powerful reasoning capabilities and extensive pre-trained knowledge. For example, GPT-Driver (\cite{mao2023gptdriver}) pioneers the reformulation of motion planning as a language modeling problem, demonstrating that LLMs can generate safe driving trajectories through natural language descriptions of coordinate positions while providing interpretable reasoning. DriveGPT4 (\cite{xu2023drivegpt4}) extends this paradigm by processing multi-frame video inputs and textual queries, which enables vehicles to interpret actions and answer user questions. DriveLM (\cite{sima2023drivelm}) introduces graph-structured visual question answering to connect perception, prediction, and planning tasks through human-written reasoning logic, facilitating more explainable decision-making. These approaches demonstrate promising semantic understanding and interpretability. Nevertheless, a fundamental gap remains: these methods primarily operate at the semantic reasoning and high-level planning abstraction, rather than delivering end-to-end planning or closed-loop control with continuous, high-frequency actuation commands.

To bridge this gap between semantic understanding and actionable planning, recent research has explored multimodal large language models (MLLMs)-powered autonomous driving systems that connect high-level reasoning to actionable trajectory planning and control within a unified framework. For example, LMDrive (\cite{shao2024lmdrive}) pioneered language-guided closed-loop driving by processing multi-view camera and LiDAR data alongside natural language instructions, enabling human-vehicle interaction through 64K instruction-following trajectories. EMMA (\cite{hwang2024emma}), developed by Waymo and built upon Google's Gemini foundation model, demonstrates how billion-parameter MLLMs can be adapted by representing all inputs and outputs including 3D object locations, trajectories, and road graphs as natural language text within a unified language space, achieving state-of-the-art motion planning performance. DriveVLM (\cite{tian2024drivevlm}) proposes a dual-system architecture that combines VLM-based scene understanding with traditional planning pipelines to balance reasoning capability with spatial precision. While these systems successfully unify semantic reasoning with continuous control, they predominantly focus on nominal driving scenarios and overlook explicit safety modeling and risk-aware planning. They lack dedicated mechanisms to identify and reason about rare but safety-critical long-tail events, such as sudden pedestrian incursions, aggressive cut-ins, or sensor occlusions, that are statistically underrepresented in training data yet account for the majority of real-world accidents. These limitations are particularly problematic in long-tail mixed-traffic scenarios, where safety-critical decisions must account for heterogeneous agents, social interactions, and rare but high-impact events.

To address these critical safety and long-tail challenges, this study introduces HERMES, an MLLM-based embodied framework that explicitly integrates safety-critical reasoning and risk-aware trajectory planning. As shown in Figure~\ref{fig:intro}, HERMES bridges the gap between end-to-end actionable planning and semantic understanding by leveraging foundation model intelligence to identify and reason particularly about rare corner cases in safety-critical scenarios. The key contributions are summarized as follows:

\begin{itemize}

\item We propose HERMES, a holistic framework that leverages large foundation models to systematically address safety-critical long-tail scenarios in end-to-end autonomous driving, where the foundation model is specifically applied to identify and reason about rare corner cases that are critical for safe trajectory planning and vehicle operation. 

\item We construct a specialized dataset for long-tail scenario reasoning based on WOD-E2E (\cite{xu2025wod-e2e}), a curated dataset for long-tail end-to-end driving. On top of WOD-E2E, we add two types of fixed offline annotations: the \textit{Long-Tail Scene Context} for multi-view hazard-centric descriptions, and the \textit{Long-Tail Planning Context} including the risk level, driving intention, high-level directives, and planning rationale, which provides an explicit foundation for risk-informed planning and safety-critical decision-making in long-tail scenarios.

\item We introduce a multimodal fusion architecture, termed the \textit{Tri-Modal Driving Module}. It delivers risk-aware and accurate end-to-end trajectory planning through the fusion of multi-view camera features, historical ego-motion patterns, and long-tail instruction embeddings encoded from the long-tail contexts.


\item We conduct comprehensive evaluations on the WOD-E2E benchmark, which shows that HERMES achieves the best performance among the state-of-the-art methods. Controlled ablation studies further validate the contributions of the key components of HERMES.
\end{itemize}

\section{Related Work}

\subsection{Foundation-Model-Driven End-to-End Driving}
Recent years have witnessed rapid progress in end-to-end autonomous driving driven by large foundation models, where planning is increasingly cast as a unified sequence modeling problem over multimodal observations and structured outputs. A representative line of work reformulates motion planning into language modeling, prompting and fine-tuning large language models to generate discretized trajectory tokens while exposing interpretable reasoning traces (\cite{mao2023gptdriver, gan2025planning}). Beyond pure text generation, multimodal LLM and VLM systems have been developed to jointly process video or multi-view visual inputs and produce driving actions together with natural-language explanations, improving transparency and human interaction (\cite{xu2023drivegpt4,sima2023drivelm, you2026v2x, you2025v2x, huang2025vlm}). Toward more deployable autonomy, LMDrive (\cite{shao2024lmdrive}) introduces a language-guided, closed-loop end-to-end driving framework that integrates multimodal sensor inputs with textual navigation instructions in interactive environments. In parallel, EMMA (\cite{hwang2024emma}) advances a generalist MLLM that unifies planning, perception, and map-related outputs within a language-centric interface, demonstrating the potential of foundation models for multi-task driving. More recent efforts further examine efficiency and systematization, such as OpenEMMA (\cite{xing2025openemma}) and LightEMMA (\cite{qiao2025lightemma}), which provide lightweight or open implementations to facilitate reproducible evaluation of VLM-based driving agents under practical constraints. Vision--Language--Action (VLA) models also emerge as a closely related paradigm that directly outputs actions or trajectory tokens from multimodal inputs, such as OpenDriveVLA (\cite{zhou2025opendrivevla}) and AutoVLA (\cite{zhou2025autovla}). 

Despite these advances, existing foundation-model-powered end-to-end driving methods still mainly focus on general semantic understanding and nominal planning, with limited dedicated design for safety-critical long-tail events and risk-aware trajectory generation in complex and mixed traffic environments.

\subsection{Long-Tail Modeling in Autonomous Driving}
Long-tail challenges in autonomous driving manifest in many forms, and prior work has studied a range of representative regimes rather than a fixed taxonomy. For low-illumination conditions, Dark Model Adaptation (\cite{dai2018dark}) targets the day-to-night domain gap for driving-scene understanding, and GCMA (\cite{sakaridis2019gcma}) further addresses nighttime ambiguity via a progressive day--twilight--night adaptation curriculum with uncertainty-aware evaluation. For adverse weather and degraded visibility, ACDC (\cite{sakaridis2021acdc}) provides a systematic benchmark covering fog, rain, snow, and night with correspondences to support controlled robustness evaluation, while Seeing Through Fog Without Seeing Fog (\cite{bijelic2020fog}) demonstrates that adaptive multimodal fusion can generalize to unseen harsh weather without explicit fog estimation. For work zones with temporary topology and rule changes, ROADWork (\cite{ghosh2025roadwork}) highlights work zones as a distinct long-tail regime and shows that strong foundation models underperform without targeted adaptation. Beyond condition-specific robustness, long-tail safety is also approached through knowledge transfer and generation: DiMA (\cite{hegdediMA2025}) distills multimodal large-model knowledge to improve planning robustness to rare events while enabling efficient inference, and generative world-model efforts such as GAIA-1 (\cite{hu2023gaia1}) and DriveDreamer-2 (\cite{zhao2024drivedreamer2}) enable controllable synthesis of rare hazards for scalable stress testing and data augmentation. Recent surveys further systematize world-model-based simulation as an emerging backbone for long-tail coverage and evaluation under distribution shifts (\cite{feng2025worldmodels_survey}). Latest industry-scale efforts further highlight reasoning-augmented VLA models, and for instance, Alpamayo-R1 (\cite{wang2025alpamayoR1}) couples interpretable chain-of-causation reasoning with action prediction and reports improved generalization under long-tail conditions.

Nevertheless, prior work often addresses scenario-specific robustness on dedicated driving functions or data synthesis in isolation, leaving a broader gap in systematically integrating long-tail signals into end-to-end trajectory planning with explicit and controllable safety preference, particularly in mixed-traffic environments involving heterogeneous road users.

\section{Methodology}
\subsection{Overview of HERMES}
\label{sec:overview}

HERMES addresses safety-critical long-tail scenarios in autonomous driving by incorporating structured risk-aware semantic guidance derived from large VLMs into an efficient end-to-end trajectory planning system. These long-tail scenarios primarily arise in mixed-traffic environments involving heterogeneous road users, where complex interactions are rare but high-impact events pose significant challenges for safe trajectory planning. Hence, the core idea of HERMES is to leverage structured semantic instructions generated by a foundation model as auxiliary guidance, which enables a lightweight model to handle rare and hazardous situations while maintaining real-time inference capability.

\begin{figure*}[]
  \centering
  \includegraphics[width=\textwidth]{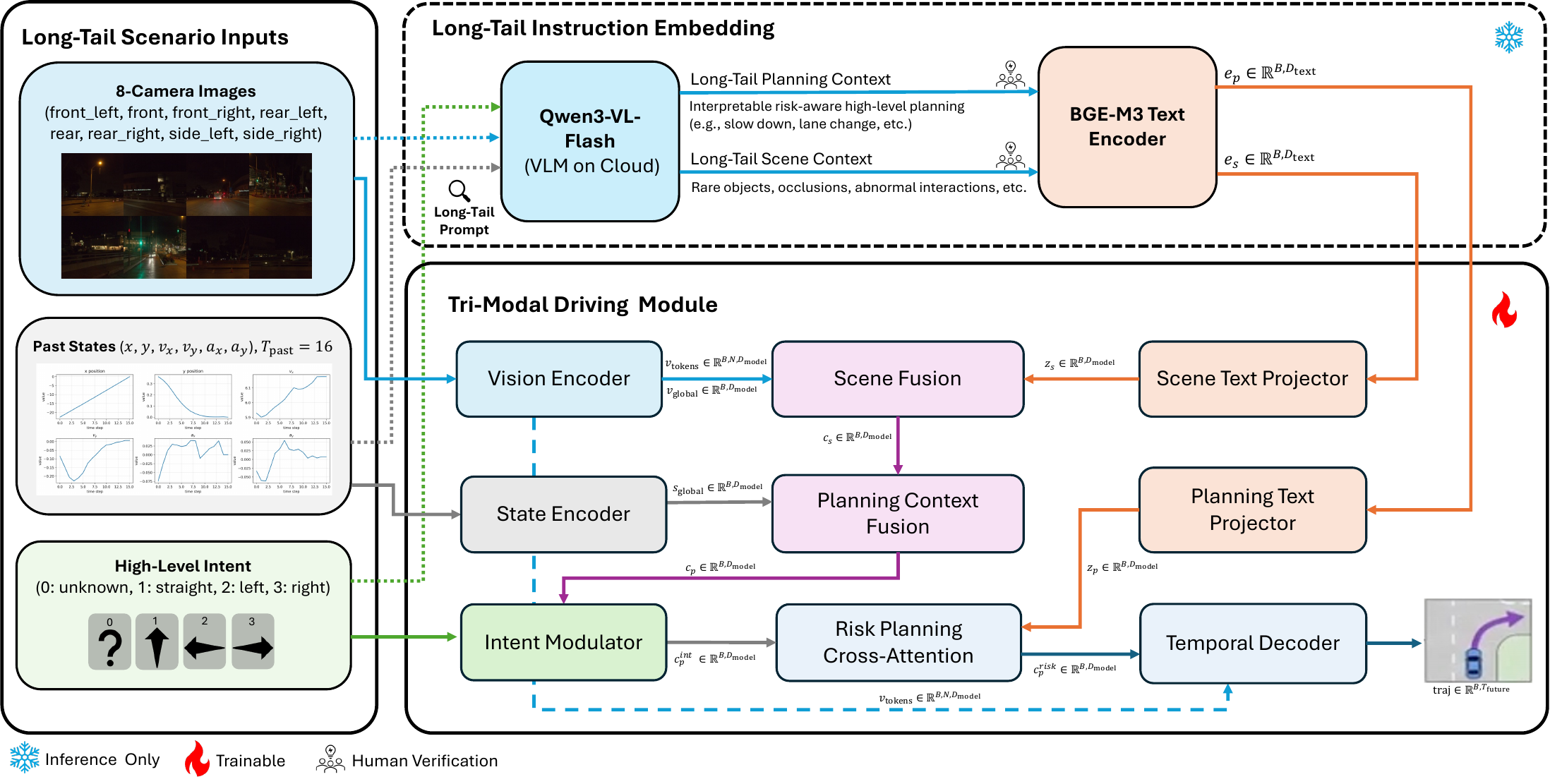}
  \caption{Overview of the HERMES framework. The \textit{Long-Tail Instruction Embedding} 
  module employs a cloud-based VLM (e.g., Qwen3-VL-Flash) to analyze 8-camera 
  images and generate structured annotations comprising \textit{Long-Tail Scene Context} and 
  \textit{Long-Tail Planning Context}, which are encoded by BGE-M3 into text embeddings. The 
  \textit{Tri-Modal Driving Module} sequentially processes multimodal 
  inputs: the \textit{Vision Encoder} first extracts spatial features, which are fused with 
  scene embeddings via \textit{Scene Fusion} to produce scene context; subsequently, the \textit{State 
  Encoder} processes historical trajectories, which are integrated with scene context 
  through \textit{Planning Context Fusion}; finally, \textit{Intent Modulator}, \textit{Risk Planning Cross-Attention}, 
  and \textit{Temporal Decoder} generate risk-aware trajectories by balancing VLM guidance with 
  learned trajectory patterns.}
  \label{fig:overview}
\end{figure*}

As illustrated in Figure~\ref{fig:overview}, HERMES consists of two main components: a \textit{Long-Tail Instruction Embedding} module, which functions to provide semantic guidance, and a \textit{Tri-Modal Driving Module}, which leverages this guidance to perform real-time trajectory planning. The \textit{Long-Tail Instruction Embedding} module employs a cloud-based VLM, to analyze multi-view surround images together with vehicle motion history, producing structured semantic instructions in the form of \textit{Long-Tail Scene Context} and \textit{Long-Tail Planning Context}. The \textit{Long-Tail Scene Context} describes rare and safety-critical elements in the driving environment such as occlusions, abnormal interactions, and uncommon objects from multiple viewpoints, while the \textit{Long-Tail Planning Context} provides interpretable risk-aware high-level planning guidance based on the identified hazards. In our experiments, these instructions are pre-computed and annotated to facilitate efficient training and reproducibility.

The \textit{Tri-Modal Driving Module} performs real-time trajectory planning by jointly processing visual observations, historical vehicle states, high-level driving intent, and the pre-computed instructions. Specifically, multi-view visual observations are first encoded to obtain spatial representations of the surrounding environment. These visual features are then augmented with the encoded \textit{Long-Tail Scene Context} to produce scene-aware representations that emphasize rare and safety-critical elements. In parallel, historical vehicle states are encoded to summarize temporal motion patterns. The resulting visual and temporal representations are subsequently fused to form a unified planning context, which is further modulated by high-level driving intent and refined using the \textit{Long-Tail Planning Context}. Through this process, the final planning representation integrates rich information from local observations together with VLM-derived long-tail contexts, enabling accurate and safety-aware vehicle operation.

Formally, given multi-view camera observations $\mathcal{I}$, a historical ego-state sequence $\mathcal{S}$, a high-level intent $\mathit{intent}$, and the two pre-computed semantic embeddings $e_{s}$ and $e_{p}$, the trajectory-planning process is formulated as
\begin{equation}
\text{traj}
=
F_{\theta}\!\left(
\mathcal{I},\mathcal{S},\mathit{intent},
e_{s},e_{p}
\right),
\label{eq:overall_formulation}
\end{equation}
where $F_{\theta}$ denotes the trainable Tri-Modal Driving Module and $\text{traj}$ represents the predicted future ego trajectory.


\subsection{Prompt Crafting for Long-Tail Annotation}
\label{sec:prompt}

The quality of VLM-generated annotations critically depends on carefully designed prompts that elicit structured, consistent, and safety-focused reasoning. Unlike general-purpose vision-language tasks, safe and reliable autonomous driving requires the model to produce actionable planning directives grounded in precise spatial understanding and risk assessment. 
To address this, we develop a specialized prompt template tailored for annotating the WOD-E2E dataset (\cite{xu2025wod-e2e}), which focuses on long-tail and safety-critical driving scenarios. The prompt explicitly guides the VLM to analyze such scenarios through a structured reasoning process, enabling the generation of high-quality long-tail annotations suitable for end-to-end driving research.

\begin{table*}[t]
\centering
\caption{Structured VLM prompt design for long-tail scenario annotation}
\label{tab:prompt}
\small
\begin{tabular}{m{0.18\textwidth}m{0.76\textwidth}}
\toprule
\textbf{Component} & \textbf{Prompt Content} \\ 
\midrule

\multicolumn{2}{l}{\textit{Input Specification}} \\
\midrule
\textbf{System Role} & 
You are an autonomous driving planner specializing in safety-critical long-tail scenario analysis. Your task is to identify rare hazardous elements and generate risk-aware planning strategies. \\ \hline

\textbf{Camera Images} & 
You are given four grouped image inputs with fixed ordering:
Image 1 contains FRONT\_LEFT, FRONT, and FRONT\_RIGHT views;
Image 2 contains REAR\_LEFT, REAR, and REAR\_RIGHT views;
Image 3 contains the SIDE\_LEFT view;
Image 4 contains the SIDE\_RIGHT view.
The ego-centric coordinate frame follows the convention that the forward direction is positive and the left direction is positive. \\ \hline 

\textbf{Motion States} & 
You are provided with historical ego vehicle motion states expressed in the same
ego-centric coordinate frame as the camera inputs, including past positions
[x, y], velocities [v\_x, v\_y], and accelerations [a\_x, a\_y] over multiple timesteps. A high-level driving intent is also provided as one of {unknown, go straight, turn left, turn right}. Use this information to infer motion trends and detect anomalous behaviors. \\ \hline

\midrule
\multicolumn{2}{l}{\textit{Long-Tail Scene Context}} \\
\midrule
\textbf{Scene Description} & 
Describe the driving scene using a multi-view structure organized by viewpoint. Report observations separately for the front, rear, side-left, and side-right views. Focus on objective, image-grounded descriptions of the scene. \\ 

& 
Include basic traffic elements such as road layout, lanes, vehicles, pedestrians, cyclists, obstacles, buildings, and intersections. \\

& 
Pay special attention to long-tail hazards, including occlusions, abnormal agent behaviors, rare objects, adverse environmental conditions, and compound edge cases involving multiple simultaneous risks. \\ \hline

\midrule
\multicolumn{2}{l}{\textit{Long-Tail Planning Context}} \\
\midrule
\textbf{Risk Level} & 
Assess the overall safety risk of the scenario and classify it as low, medium, or high based on scene complexity, vulnerable road users, collision likelihood, and reaction time. \\ \hline

\textbf{Intention} & 
Infer the high-level driving maneuver for the ego vehicle.
Choose one action from go straight, turn left, turn right, or stop. \\ \hline

\textbf{High-Level Planning} & 
Provide a concise planning directive describing the intended driving behavior, such as maintaining lane, yielding, braking, or proceeding cautiously. \\ \hline

\textbf{Planning Rationale} & 
Briefly explain the reasoning behind the selected plan.
The explanation should be grounded in observable scene elements or ego vehicle dynamics and should not introduce unobserved information. \\ \hline

\midrule
\multicolumn{2}{l}{\textit{Output Constraint}} \\
\midrule
\textbf{Output Format} & 
Produce exactly five output fields in the following order: [scene description], [risk level], [trajectory intention], [high-level plan], and [plan rationale]. No additional information should be included beyond these fields. \\
\bottomrule
\end{tabular}
\end{table*}

As summarized in Table~\ref{tab:prompt}, our prompt design is structured to guide the VLM to act as an autonomous driving planner specialized in safety-critical long-tail scenario analysis. The prompt explicitly instructs the model to identify rare hazardous elements from multi-view observations and to generate risk-aware planning strategies grounded in both perception and motion context.

The prompt specification consists of a fixed input description and structured semantic outputs corresponding to long-tail scene understanding and planning guidance. For the input, the model is provided with multi-view surround images arranged into four grouped inputs with a fixed ordering, together with an ego-centric coordinate frame in which the forward and left directions are consistently defined. This explicit specification reduces viewpoint ambiguity and helps prevent incorrect spatial interpretation. In addition, historical ego vehicle motion states are supplied in the same coordinate frame, including past positions, velocities, and accelerations over multiple timesteps, along with a high-level driving intent, enabling the model to reason about temporal motion patterns and dynamic behaviors.

Based on these inputs, the prompt elicits two categories of semantic outputs. The first corresponds to the \textit{Long-Tail Scene Context}, which requires the model to produce a
multi-view description of the driving scene organized by viewpoint, including front, rear, side-left, and side-right observations. The scene description covers both basic traffic elements and long-tail hazards, such as occlusions, abnormal agent behaviors, rare objects, adverse environmental conditions, and compound edge cases involving
multiple simultaneous risks. This structured representation ensures comprehensive coverage of spatially distributed hazards that may only be visible from specific perspectives.

The second category corresponds to the \textit{Long-Tail Planning Context}, which captures decision-level semantic guidance. Specifically, the model is required to assess the
overall safety risk of the scenario, infer a high-level maneuver intention, generate a concise planning directive, and provide a brief rationale grounded in observable scene
elements or ego vehicle dynamics. Together, these outputs encode risk-aware planning knowledge that complements geometric trajectory supervision.

Finally, all outputs are constrained to follow a fixed structured format consisting of exactly five fields, ensuring consistency across annotations and reliable downstream integration. This prompt design forms the foundation of our long-tail instruction dataset, which enables the generation of high-quality annotations that capture both
perceptual context and planning intent. These annotations are subsequently encoded into text embeddings and used to guide the trajectory planning model during training and inference, allowing VLM-derived semantic reasoning to be incorporated into an efficient end-to-end planning framework.

\subsection{Long-Tail Instruction Embedding}
\label{sec:embedding}

The \textit{Long-Tail Instruction Embedding} module provides semantic supervision that guides the \textit{Tri-Modal Driving Module} in safety-critical and long-tail driving scenarios. As illustrated in Figure~\ref{fig:overview}, using the crafted prompt template described in the above section, this module leverages a cloud-based VLM to analyze 8-camera surround-view images together with historical vehicle motion states and high-level intent, producing structured semantic instruction annotations that capture both environmental hazards and planning considerations.

The generated textual annotations, comprising the \textit{Long-Tail Scene Context} and \textit{Long-Tail Planning Context}, are subsequently encoded into fixed-dimensional vector representations using BGE-M3 (\cite{chen2024bge}), a strong and efficient sentence embedding model that produces semantically discriminative representations. Specifically, the \textit{Long-Tail Scene Context} is encoded as a scene embedding denoted as $e_{s}$ $ \in \mathbb{R}^{B \times D_{\text{text}}}$, while the \textit{Long-Tail Planning Context} is encoded as a planning embedding denoted as $e_{p}$ $\in \mathbb{R}^{B \times D_{\text{text}}}$, where $B$ is the batch size and $D_{\text{text}}$ is the embedding dimension of the text encoder. 


Both text embeddings serve as compact semantic representations of long-tail knowledge and are treated as fixed inputs to the \textit{Tri-Modal Driving Module}. As shown in Figure~\ref{fig:overview}, $e_{s}$ and $e_{p}$ are passed to dedicated text projection modules in the \textit{Tri-Modal Driving Module}, where they are mapped into the model’s latent space and fused with visual and temporal features for downstream trajectory planning.

\subsection{Tri-Modal Driving Module}
\label{sec:trimodal}

The \textit{Tri-Modal Driving Module} performs real-time trajectory planning by integrating visual, temporal, and linguistic information under semantic guidance from the \textit{Long-Tail Instruction Embedding} module. As illustrated in Figure~\ref{fig:overview}, 
this module processes multi-view camera images, historical vehicle states, 
high-level driving intent, and pre-computed long-tail instruction embeddings through a 
sequential pipeline of specialized components, ultimately generating safe and 
risk-informed trajectories. The module comprises the following key components.

\textbf{Vision Encoder.}
The \textit{Vision Encoder} adopts a Vision Transformer (ViT) architecture (\cite{dosovitskiy2021image})
to extract spatial representations from the eight surround-view camera images.
For each camera $c\in\{1,\ldots,C\}$, where $C=8$, the ViT encoder produces
$N_p$ patch-level features. Let
$\mathit{v}_{c,n}^{\mathrm{vit}}\in\mathbb{R}^{D_{\mathrm{vit}}}$
denote the $n$-th patch feature extracted from camera $c$, where
$n\in\{1,\ldots,N_p\}$ and $D_{\mathrm{vit}}$ is the output feature
dimension of the ViT backbone.

To align the visual features with the common latent space used by the
subsequent multimodal modules, each patch feature is first linearly projected
to $D_{\mathrm{model}}$. Learnable camera-specific and viewpoint-specific
embeddings are then incorporated to preserve the spatial source of each
visual token:
\begin{equation}
\mathit{v}_{c,n}
=
W_{\mathrm{v}}\mathit{v}_{c,n}^{\mathrm{vit}}
+
\mathit{e}_{c}^{\mathrm{cam}}
+
\mathit{e}_{c}^{\mathrm{view}},
\label{eq:vision_token}
\end{equation}
where
$W_{\mathrm{v}}\in
\mathbb{R}^{D_{\mathrm{model}}\times D_{\mathrm{vit}}}$
denotes the visual projection matrix, and
$\mathit{e}_{c}^{\mathrm{cam}},
\mathit{e}_{c}^{\mathrm{view}}
\in\mathbb{R}^{D_{\mathrm{model}}}$
denote the learnable camera-specific embedding and the viewpoint embedding
associated with camera $c$, respectively. Consequently,
$\mathit{v}_{c,n}\in\mathbb{R}^{D_{\mathrm{model}}}$. These embeddings explicitly encode the spatial origin of each visual token,
helping the model distinguish observations from different camera directions
and preserve viewpoint-specific information during multi-view feature fusion.

The patch representations from all cameras are concatenated into
$\mathit{v_{\text{tokens}}}\in
\mathbb{R}^{B\times N\times D_{\mathrm{model}}}$,
where $B$ is the batch size and $N=C N_p$ denotes the total number of
visual tokens across all cameras. In addition to the token-level
representation, a compact global visual representation is obtained through
mean pooling:
\begin{equation}
v_{\mathrm{global}}
=
\frac{1}{C N_p}
\sum_{c=1}^{C}
\sum_{n=1}^{N_p}
v_{c,n}.
\end{equation}
where
$\mathit{v_{\text{global}}}\in
\mathbb{R}^{B\times D_{\mathrm{model}}}$.
The token-level representation $\mathit{v_{\text{tokens}}}$ preserves
fine-grained spatial information for subsequent cross-attention and temporal
decoding, while $\mathit{v_{\text{global}}}$ provides a compact summary of
the multi-view visual observations for scene-level fusion.

\textbf{State Encoder.}
The \textit{State Encoder} captures the temporal dynamics of the ego vehicle
from its historical motion states. Let the historical ego-state sequence be
$\mathcal{S}\in
\mathbb{R}^{B\times T_{\mathrm{past}}\times D_{\mathrm{state}}}$,
where each
$\mathit{s}_{t}\in\mathbb{R}^{D_{\mathrm{state}}}$
denotes the ego state at timestep $t$. In our implementation,
$D_{\mathrm{state}}=6$, with
$\mathit{s}_{t}=[x_t,y_t,v_{x,t},v_{y,t},a_{x,t},a_{y,t}]$,
representing the ego position, velocity, and acceleration.

Each state vector is first mapped from the motion-state space to the common
latent space through a learnable linear projection:
\begin{equation}
\mathit{h}_{t}^{(0)}
=
W_{\mathrm{s}}\mathit{s}_{t}
+
\mathit{b}_{\mathrm{s}},
\end{equation}
where
$W_{\mathrm{s}}\in
\mathbb{R}^{D_{\mathrm{model}}\times D_{\mathrm{state}}}$
and
$\mathit{b}_{\mathrm{s}}\in
\mathbb{R}^{D_{\mathrm{model}}}$
are learnable parameters, yielding
$\mathit{h}_{t}^{(0)}\in\mathbb{R}^{D_{\mathrm{model}}}$.
This projection transforms the low-dimensional motion states into the same
latent dimension as the other modalities, facilitating subsequent multimodal
feature fusion.

The projected sequence is subsequently processed by a two-layer Transformer
encoder with four attention heads:
\begin{equation}
\mathit{H}_{\mathrm{s}}
=
\mathrm{Transformer}_{\mathrm{s}}
\left(
\mathit{H}_{\mathrm{s}}^{(0)}
\right),
\end{equation}
where
$\mathit{H}_{\mathrm{s}}^{(0)}
\in\mathbb{R}^{B\times T_{\mathrm{past}}\times D_{\mathrm{model}}}$
denotes the sequence of projected state features and
$\mathit{H}_{\mathrm{s}}\in
\mathbb{R}^{B\times T_{\mathrm{past}}\times D_{\mathrm{model}}}$
denotes the temporally encoded representation.
The self-attention mechanism allows the encoder to model dependencies across
historical timesteps, thereby capturing recent ego-motion patterns beyond
independent state observations.

Finally, temporal mean pooling is applied to summarize the encoded motion
history into a compact global state representation:
\begin{equation}
\mathit{s_{\text{global}}}
=
\frac{1}{T_{\mathrm{past}}}
\sum_{t=1}^{T_{\mathrm{past}}}
\mathit{H}_{\mathrm{s},t},
\end{equation}
where
$\mathit{s_{\text{global}}}\in
\mathbb{R}^{B\times D_{\mathrm{model}}}$.
This global representation summarizes the recent ego-motion dynamics and is
subsequently combined with the scene-aware visual representation in the
\textit{Planning Context Fusion} module.

\textbf{Scene Fusion.}
The \textit{Scene Fusion} module integrates the \textit{Long-Tail Scene Context} with
multi-view visual features to construct a scene-aware representation for
subsequent trajectory planning. Its inputs consist of the semantic scene
embedding $e_{s} \in\mathbb{R}^{B\times D_{\text{text}}}$,
the token-level visual representation
$\mathit{v_{\text{tokens}}}\in
\mathbb{R}^{B\times N\times D_{\text{model}}}$,
and the global visual representation
$\mathit{v_{\text{global}}}\in
\mathbb{R}^{B\times D_{\text{model}}}$.

Since the semantic embedding and visual features are represented in different
feature spaces, the \textit{Long-Tail Scene Context} is first mapped from the text
embedding space to the common model latent space through a learnable linear
projection:
\begin{equation}
z_{s}
=
W_{\mathrm{scene}}\,
e_{s}
+
\mathit{b}_{\mathrm{scene}},
\end{equation}
where
$W_{\mathrm{scene}}\in
\mathbb{R}^{D_{\mathrm{model}}\times D_{\text{text}}}$
and
$\mathit{b}_{\mathrm{scene}}\in
\mathbb{R}^{D_{\mathrm{model}}}$
are learnable parameters, yielding
$z_{s}\in
\mathbb{R}^{B\times D_{\mathrm{model}}}$.
This projection aligns the semantic representation with the visual feature
dimension, enabling subsequent cross-modal interaction in a shared latent
space.

The projected semantic representation is then used as a query to aggregate
scene-relevant information from the multi-view visual tokens through
multi-head cross-attention:
\begin{equation}
\mathit{c}_{\text{scene}}
=
\mathrm{MHA}
\!\left(
\mathrm{LN}(z_{s}),
\mathit{v_{\text{tokens}}},
\mathit{v_{\text{tokens}}}
\right),
\end{equation}
where the three arguments correspond to the query, key, and value,
respectively, and
$\mathit{c}_{\text{scene}}\in
\mathbb{R}^{B\times D_{\text{model}}}$.
By querying the visual tokens with the semantic scene representation, the
module aggregates visual evidence associated with the long-tail scene
description rather than relying solely on a globally pooled visual feature.

The attended visual feature is subsequently integrated with the original
semantic representation through a residual connection followed by layer
normalization:
\begin{equation}
\mathit{c}_{\text{text}}
=
\mathrm{LN}
\!\left(
z_{s}
+
\mathit{c}_{\text{scene}}
\right),
\end{equation}
where
$\mathit{c}_{\text{text}}\in
\mathbb{R}^{B\times D_{\text{model}}}$.
The residual connection preserves the original semantic information while
incorporating visual evidence obtained through cross-attention, thereby
maintaining both semantic guidance and observation-grounded scene features.

The resulting cross-modal representation is then concatenated with the global
visual representation and processed by a two-layer MLP with GELU activation:
\begin{equation}
\tilde{\mathit{s}}
=
\mathrm{MLP}_{\mathrm{scene}}
\!\left(
\left[
\mathit{v_{\text{global}}}
\,\Vert\,
\mathit{c}_{\text{text}}
\right]
\right),
\end{equation}
where
$[\mathit{v_{\text{global}}} \,\Vert\, \mathit{c}_{\text{text}}]
\in \mathbb{R}^{B\times 2D_{\text{model}}}$
and
$\tilde{\mathit{s}}\in
\mathbb{R}^{B\times D_{\text{model}}}$.
This operation combines the global visual summary with the semantically
grounded visual representation, allowing the module to retain overall scene
information while emphasizing features relevant to the \textit{Long-Tail Scene
Context}.

Finally, the fused representation is added back to the global visual feature
through a residual connection:
\begin{equation}
c_{s}
=
\mathit{v_{\text{global}}}
+
\tilde{\mathit{s}},
\end{equation}
where
$c_{s}\in
\mathbb{R}^{B\times D_{\text{model}}}$.
The resulting scene-aware context therefore preserves the global visual
representation while introducing semantic guidance derived from the
\textit{Long-Tail Scene Context}, and is subsequently combined with the encoded
ego-motion representation in the \textit{Planning Context Fusion} module.

\textbf{Planning Context Fusion.}
The scene-aware representation $c_{s}$ is combined with the 
temporal state summary $\mathit{s_{\text{global}}}$ through the 
\textit{Planning Context Fusion} module, implemented as a two-layer MLP, yielding 
the planning context:
\begin{equation}
c_{p} = \text{MLP}\!\left(
\left[c_{s} \, \Vert \, \mathit{s_{\text{global}}}\right]
\right).
\end{equation}
The resulting planning context $c_{p} \in \mathbb{R}^{B \times D_{\text{model}}}$ 
provides a unified representation that integrates scene-level semantics with 
historical motion dynamics. The resulting $c_{p} \in 
\mathbb{R}^{B \times D_{\text{model}}}$ forms a motion-grounded semantic planning 
representation before maneuver-level intent modulation and risk-aware semantic 
conditioning.

\textbf{Intent Modulator.}
The \textit{Intent Modulator} incorporates high-level maneuver guidance into
the motion-grounded planning representation through context-dependent
feature-wise modulation. The high-level intent is represented by
$\mathit{intent}\in\{0,1,2,3\}^{B}$, corresponding to unknown, straight,
left, and right, respectively. A learnable embedding layer first maps each
discrete intent identifier into the common model latent space:
\begin{equation}
\mathit{e}_{\text{intent}}
=
E_{\text{intent}}(\mathit{intent}),
\end{equation}
where
$E_{\text{intent}}\in
\mathbb{R}^{4\times D_{\text{model}}}$
denotes the learnable intent embedding table and
$\mathit{e}_{\text{intent}}\in
\mathbb{R}^{B\times D_{\text{model}}}$.

Rather than directly appending the intent embedding to the downstream decoder
input, the modulator jointly considers the current planning representation
and the maneuver intent. Specifically,
$c_{p}\in
\mathbb{R}^{B\times D_{\text{model}}}$
and $\mathit{e}_{\text{intent}}$ are concatenated and normalized, and the
resulting representation is processed by a two-layer MLP with GELU activation:
\begin{equation}
\mathit{z}_{\text{intent}}
=
f_{\text{intent}}
\left(
\mathrm{LN}
\left(
[c_{p} \,\Vert\, \mathit{e}_{\text{intent}}]
\right)
\right),
\end{equation}
where
$[c_{p} \,\Vert\, \mathit{e}_{\text{intent}}]
\in\mathbb{R}^{B\times 2D_{\text{model}}}$,
$f_{\text{intent}}(\cdot)$ denotes the two-layer modulation MLP, and
$\mathit{z}_{\text{intent}}\in
\mathbb{R}^{B\times 2D_{\text{model}}}$.

The modulation output is equally divided along the feature dimension into
the scale $\gamma$ and shift $\beta$ parameters:
\begin{equation}
[\gamma \,\Vert\, \beta]
=
\mathrm{Split}(\mathit{z}_{\text{intent}}),
\end{equation}
where
$\gamma,\beta\in
\mathbb{R}^{B\times D_{\text{model}}}$.
The scale $\gamma$ parameters are bounded through a sigmoid function, and the
intent-aware planning representation is computed through a feature-wise
affine transformation:
\begin{equation}
c^{int}_{p}
=
c_{p}
\odot
\left(
1+\sigma(\gamma)
\right)
+
\beta,
\end{equation}
where $\sigma(\cdot)$ denotes the sigmoid function,
$\odot$ denotes element-wise multiplication, and
$c^{int}_{p}\in
\mathbb{R}^{B\times D_{\text{model}}}$.

This design differs from direct intent concatenation in that the intent
embedding is not simply provided as an additional downstream feature.
Instead, the current planning context and maneuver intent jointly determine
how individual dimensions of the planning representation are rescaled and
shifted. The modulation is therefore context-dependent: the same maneuver
intent can produce different feature transformations under different visual
and ego-motion contexts. This design preserves the motion-grounded planning
representation as the primary feature while allowing high-level maneuver
information to explicitly adjust its feature-wise response.

For cases in which a reliable maneuver intent is unavailable, the
\textit{unknown} category is represented by its own learnable embedding in
$E_{\text{intent}}$. It is processed through the same modulation mechanism
rather than being manually mapped to any specific maneuver. Consequently,
the model can learn the modulation associated with unspecified intent
directly from the training data without imposing a predefined straight,
left, or right maneuver assumption.

\begin{figure}[]
  \centering
  \includegraphics[width=0.7\textwidth]{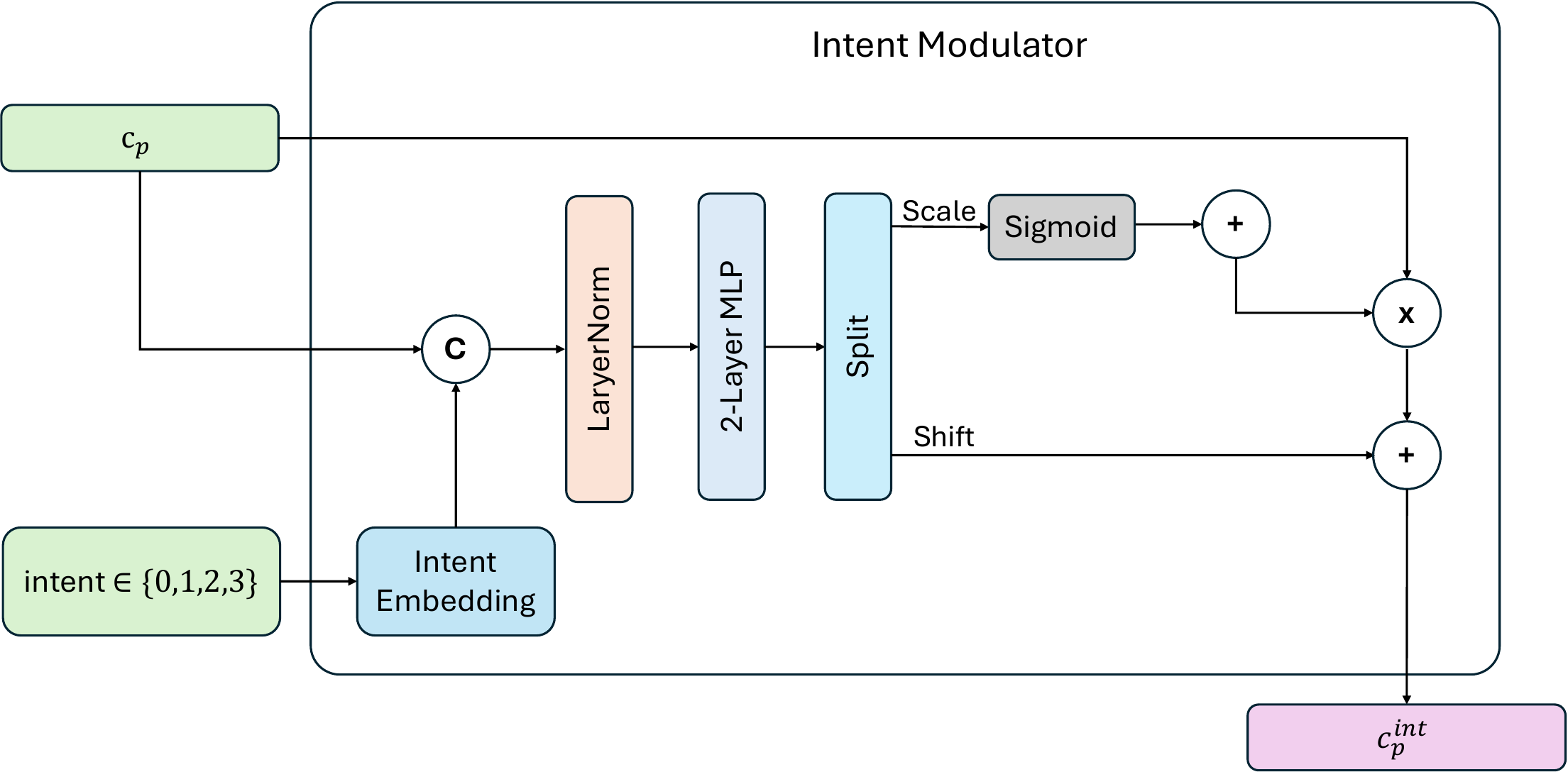}
  \caption{Architecture of the \textit{Intent Modulator}. High-level driving intent 
  is embedded and processed through a two-layer MLP to generate scale $\gamma$ and shift $\beta$
  parameters, which adaptively modulate the planning context.}
  \label{fig:intent_modulator}
\end{figure}

\textbf{Risk Planning Cross-Attention.}
The \textit{Risk Planning Cross-Attention} module incorporates the
\textit{Long-Tail Planning Context} as an auxiliary semantic conditioning signal for
trajectory planning. The input
$e_{p}\in
\mathbb{R}^{B\times D_{\text{text}}}$
contains the semantic representation of the long-tail risk assessment and
planning guidance. It is first projected into the common model latent space:
\begin{equation}
z_{p}
=
W_{\mathrm{risk}}\,
e_{p}
+
\mathit{b}_{\mathrm{risk}},
\end{equation}
where
$W_{\mathrm{risk}}\in
\mathbb{R}^{D_{\mathrm{model}}\times D_{\text{text}}}$
and
$\mathit{b}_{\mathrm{risk}}\in
\mathbb{R}^{D_{\mathrm{model}}}$
are learnable parameters, yielding
$z_{p}\in
\mathbb{R}^{B\times D_{\mathrm{model}}}$.
The projection aligns the semantic planning representation with the latent
dimension of the trajectory-planning features.

The intent-aware planning representation is then used as the query, while
the projected \textit{Long-Tail Planning Context} serves as the key and value:
\begin{equation}
\mathit{c}_{\mathrm{risk}}
=
\mathrm{MHA}
\left(
c^{int}_{p},
z_{p},
z_{p}
\right),
\end{equation}
where
$c^{int}_{p}\in
\mathbb{R}^{B\times D_{\mathrm{model}}}$
and
$\mathit{c}_{\mathrm{risk}}\in
\mathbb{R}^{B\times D_{\mathrm{model}}}$.
Since the \textit{Long-Tail Planning Context} is represented by a single semantic
vector, this operation is not intended to perform token-level retrieval over
multiple planning instructions. Instead, it transforms the semantic planning
signal into the model latent space for subsequent residual conditioning of
the motion-grounded planning representation.

The semantic adjustment is incorporated through a weighted residual
connection followed by layer normalization:
\begin{equation}
c_{p}^{risk}
=
\mathrm{LN}
\left(
c_{p}^{int}
+
\alpha\,\mathit{c}_{\mathrm{risk}}
\right),
\end{equation}
where $\alpha$ is a residual weighting coefficient and
$c_{p}^{risk}\in
\mathbb{R}^{B\times D_{\mathrm{model}}}$.
In our implementation, $\alpha$ is set to $0.3$, such that the
motion-grounded and intent-conditioned planning representation remains the
primary information pathway, while the \textit{Long-Tail Planning Context} provides
a controlled semantic adjustment.

This controlled residual design is motivated by the different reliability
characteristics of the two information sources. During dataset construction in this study, all generated semantic annotations are manually verified for training and validation to ensure efficient learning.
However, in a potential online deployment, the \textit{Long-Tail Planning Context}
would be generated from a foundation model directly and may therefore contain
uncertain or imperfect semantic guidance. Directly allowing such external
semantic information to dominate the planning representation could make the
trajectory planner overly dependent on its reliability. The weighted
residual formulation instead treats the semantic planning context as an
auxiliary guidance signal and limits its contribution relative to the
motion-grounded planning representation. This design is intended to reduce
over-reliance on potentially imperfect semantic guidance while retaining its
ability to adjust trajectory planning in long-tail scenarios.

\begin{figure}[]
  \centering
  \includegraphics[width=0.7\textwidth]{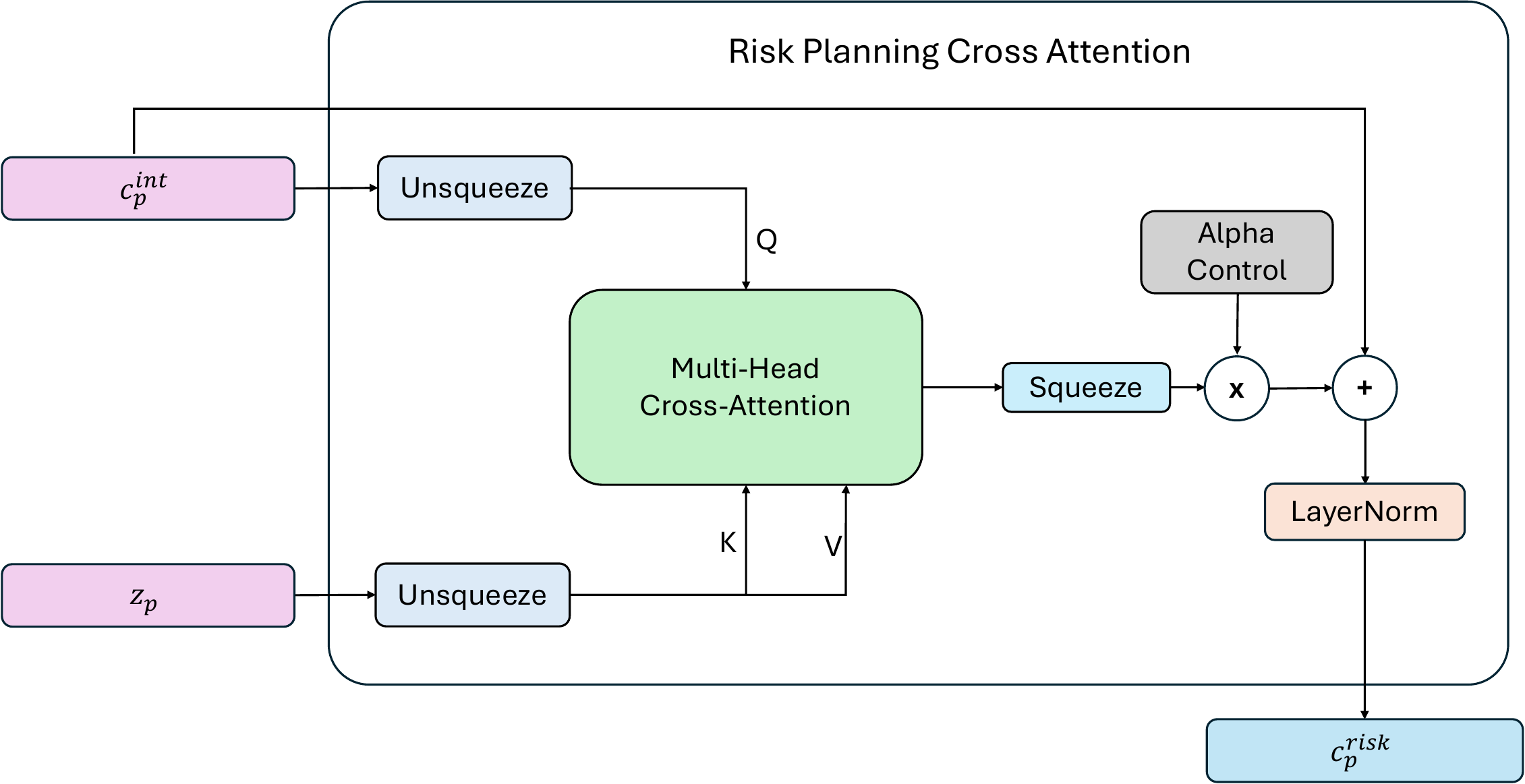}
  \caption{Architecture of the \textit{Risk Planning Cross-Attention} module. 
  The intent-aware planning context attends to the planning instruction embedding 
  through multi-head cross-attention, and the resulting semantic adjustment is 
  integrated via a scaled residual connection followed by layer normalization.}
  \label{fig:risk_attention}
\end{figure}

\textbf{Temporal Decoder.}
The \textit{Temporal Decoder} generates the future ego trajectory from the
controlled planning representation using a set of learnable time-indexed
queries. For each future timestep
$t\in\{1,\ldots,T_{\text{future}}\}$, a learnable temporal embedding is
conditioned on the planning representation:
\begin{equation}
\mathit{q}_{t}
=
\mathit{e}_{t}^{\mathrm{time}}
+
W_{\mathrm{q}}\,
c_{p}^{risk},
\end{equation}
where
$\mathit{e}_{t}^{\mathrm{time}}\in\mathbb{R}^{D_{\mathrm{model}}}$
denotes the learnable embedding associated with future timestep $t$,
$W_{\mathrm{q}}\in
\mathbb{R}^{D_{\mathrm{model}}\times D_{\mathrm{model}}}$
is a learnable query projection, and
$\mathit{q}_{t}\in\mathbb{R}^{B\times D_{\mathrm{model}}}$.
The time-indexed embeddings allow different future timesteps to maintain
distinct temporal representations, while conditioning each query on
$c_{p}^{risk}$ provides a shared planning context for the
entire prediction horizon.

In parallel, the controlled planning representation is projected to construct
a compact planning memory:
\begin{equation}
m^{(0)}
=
W_{\mathrm{m}}\,
c_{p}^{risk}
+
\mathit{b}_{\mathrm{m}},
\end{equation}
where
$W_{\mathrm{m}}\in
\mathbb{R}^{D_{\mathrm{model}}\times D_{\mathrm{model}}}$
and
$\mathit{b}_{\mathrm{m}}\in
\mathbb{R}^{D_{\mathrm{model}}}$
are learnable parameters, yielding
$m^{(0)}\in
\mathbb{R}^{B\times D_{\mathrm{model}}}$.

To retain fine-grained visual information for trajectory generation,
$m^{(0)}$ is treated as a single planning token and concatenated with the
multi-view visual tokens
$v_{tokens}\in\mathbb{R}^{B\times N\times D_{\mathrm{model}}}$
along the token dimension. The resulting decoder memory is
\begin{equation}
m
=
\mathrm{LN}
\left(
[
m^{(0)} \,\Vert\, v_{tokens}
]
\right),
\end{equation}
where the planning representation $m^{(0)}$ is expanded with a singleton token
dimension before concatenation, and
\[
m\in
\mathbb{R}^{B\times (N+1)\times D_{\mathrm{model}}}.
\]
This memory allows the decoder to use both the compact planning representation
and the token-level multi-view visual features during future trajectory
generation.

The conditioned temporal queries are then processed by a three-layer
Transformer decoder with eight attention heads:
\begin{equation}
\mathit{h}_{t}^{\mathrm{dec}}
=
\mathrm{Decoder}
\left(
\mathit{q}_{t},
m
\right),
\end{equation}
where
$\mathit{h}_{t}^{\mathrm{dec}}\in
\mathbb{R}^{B\times D_{\mathrm{model}}}$.
Each decoder output is projected to a two-dimensional relative displacement:
\begin{equation}
\Delta\mathit{p}_{t}
=
W_{\mathrm{o}}\,
\mathit{h}_{t}^{\mathrm{dec}}
+
\mathit{b}_{\mathrm{o}},
\end{equation}
where
$W_{\mathrm{o}}\in
\mathbb{R}^{2\times D_{\mathrm{model}}}$,
$\mathit{b}_{\mathrm{o}}\in\mathbb{R}^{2}$, and
$\Delta\mathit{p}_{t}\in\mathbb{R}^{B\times2}$.

The predicted relative displacements are accumulated to recover the future
ego positions:
\begin{equation}
\mathit{p}_{t}
=
\mathit{p}_{0}
+
\sum_{k=1}^{t}
\Delta\mathit{p}_{k},
\end{equation}
where
$\mathit{p}_{0}\in\mathbb{R}^{B\times2}$
denotes the initial ego position and
$\mathit{p}_{t}\in\mathbb{R}^{B\times2}$
denotes the accumulated position at future timestep $t$.
Predicting relative displacements before accumulation encourages the decoder
to model the sequential evolution of future ego motion rather than directly
regressing each future position independently.

Finally, a lightweight refinement MLP provides a residual correction to each
accumulated position:
\begin{equation}
\hat{\mathit{p}}_{t}
=
\mathit{p}_{t}
+
f_{\mathrm{ref}}
\left(
\mathit{p}_{t}
\right),
\end{equation}
where
$f_{\mathrm{ref}}:\mathbb{R}^{2}\rightarrow\mathbb{R}^{2}$
denotes the refinement network and
$\hat{\mathit{p}}_{t}\in\mathbb{R}^{B\times2}$.
The final predicted trajectory is therefore represented as
\begin{equation}
\text{traj}
=
\left[
\hat{\mathit{p}}_{1},
\ldots,
\hat{\mathit{p}}_{T_{\text{future}}}
\right]
\in
\mathbb{R}^{B\times T_{\text{future}}\times2}.
\end{equation}

It is important to note that the semantic planning context does not serve as
a fixed control command applied independently at each future timestep.
Instead, it conditions the current latent planning representation, from which
the Temporal Decoder predicts the future trajectory using distinct learnable
time-indexed queries. Therefore, the 5-second trajectory represents a
conditioned prediction based on the current observation and semantic context,
rather than the repeated execution of an unchanged semantic command over the
entire prediction horizon.

\section{Experiment}
\subsection{Dataset}
\label{sec:dataset}

We build our long-tail scenario annotations on top of the WOD-E2E dataset(\cite{xu2025wod-e2e}), a latest large-scale benchmark for learning-based autonomous driving under realistic and safety-critical conditions, and evaluate the proposed HERMES on the annotated benchmark. WOD-E2E contains 4,021 driving segments collected from diverse urban and suburban environments, covering approximately 12 hours of real-world driving. The official training, validation, and testing split consists of 2,037, 479, and 1,505 segments, respectively. The same long-tail annotation pipeline is applied to the training and validation splits. WOD-E2E emphasizes rare and safety-critical long-tail scenarios such as vulnerable road users, occlusions, unusual agent behaviors, and other challenging conditions that are underrepresented in standard driving datasets. This focus makes it a strong foundation for long-tail analysis and modeling in end-to-end autonomous driving, especially under complex and mixed traffic conditions.


\subsection{Implementation Details}
\label{sec:implementation}

We implement HERMES in PyTorch and train it on the official training split using a single NVIDIA RTX 4090 GPU with a batch size of 16. The model is trained for 3 epochs. The training objective is the Mean Squared Error (MSE) between the predicted and ground-truth future ego trajectories over a 5-second horizon, corresponding to 20 future timesteps. We use the Adam optimizer with an initial learning rate of $1\times10^{-4}$. Training employs gradient clipping with an $\ell_2$ norm threshold of 5.0, along with a linear warmup schedule over the first epoch starting from $1\times10^{-5}$, followed by cosine annealing to a minimum learning rate of $5\times10^{-6}$. Within the \textit{Risk Planning Cross-Attention} module, the control parameter $\alpha$ is set to 0.3 in all experiments.

Input images are resized to $224 \times 224$ and normalized using ImageNet statistics. Qwen3-VL-Flash (\cite{Qwen3-VL}) is applied for long-tail reasoning and annotation. Textual instruction embeddings are obtained through the BGE-M3 text encode (\cite{chen2024bge}). All components of the \textit{Tri-Modal Driving Module}, including the vision backbone and downstream fusion and decoding modules, are trained end-to-end.

\subsection{Baselines}
\label{sec:baselines}

We compare HERMES with four representative and publicly documented
foundation-model-driven end-to-end driving methods:
OpenEMMA~\citep{xing2025openemma},
LightEMMA~\citep{qiao2025lightemma},
AutoVLA~\citep{zhou2025autovla}, and
Dream-VLA~\citep{ye2025dreamvla}.
These methods cover representative VLM- and VLA-based end-to-end driving
paradigms and provide relevant comparisons with recent multimodal planning
approaches.

For these methods, we report their publicly available results on the
official WOD-E2E test set from the benchmark leaderboard
~\citep{waymo2025e2eleaderboard}. HERMES is evaluated through the same
official submission protocol and evaluation server. Accordingly, the
results reported in Tables~\ref{tab:overall_results} and
~\ref{tab:category_results} are obtained under the same official test-set
evaluation protocol and metrics. The reported OpenEMMA and Dream-VLA
results correspond to their official submissions based on QwenVLM and
Dream-VLA v3.4 GRPO, respectively. Notably, the official test set is evaluated through server-side submission
and is not available for manual inspection. Therefore, the long-tail semantic
guidance used for test-time prediction cannot be manually verified and must
be handled by the proposed controlled semantic conditioning mechanism described in Section~\ref{sec:trimodal}. This way, in turn, provides a more realistic evaluation setting that reflects real-time deployment without human intervention.




\subsection{Metrics}
\label{sec:metrics}

We evaluate end-to-end driving performance using the Rater Feedback
Score (RFS)~\citep{xu2025wod-e2e} together with standard trajectory
distance metrics, including Average Displacement Error (ADE) and Final
Displacement Error (FDE)~\citep{gupta2018socialgan}. RFS is the primary
evaluation metric of WOD-E2E and measures the quality of predicted
driving behavior using a learned evaluator derived from human preference
feedback. In contrast, ADE and FDE quantify the
geometric deviation between the predicted and ground-truth trajectories,
providing complementary measures of trajectory accuracy.

For the official WOD-E2E test-set comparison, we report RFS and ADE at 3~s and 5~s, as provided by the official evaluation server. The category-wise analysis further
reports RFS across the official long-tail scenario categories. For the ablation studies, we use the WOD-E2E validation set, which allows a more efficient and controlled evaluation of individual components under the same data split and experimental setting. We also report FDE at 3~s and 5~s for a more detailed assessment.

\subsection{Performance Evaluation}

\subsubsection{Overall Results}

Table~\ref{tab:overall_results} summarizes the official WOD-E2E test-set
performance of HERMES and the selected comparison methods. HERMES achieves
the best overall performance across all reported metrics, with an RFS of
7.3915 and ADE values of 1.5949 and 3.3060 at the 3~s and 5~s horizons,
respectively. Compared with the strongest competing results in the table,
HERMES improves RFS over Dream-VLA while also achieving lower trajectory
errors than AutoVLA, which obtains the best ADE performance among the
baseline methods.

These results demonstrate that HERMES improves both human-preference-aligned
driving quality, as measured by RFS, and geometric trajectory accuracy.
The consistent improvements across these complementary metrics indicate the
effectiveness of incorporating long-tail semantic guidance and risk-aware
conditioning into end-to-end trajectory planning.

\subsubsection{Category-wise RFS Analysis}

Table~\ref{tab:category_results} further reports RFS across the eleven
official WOD-E2E scenario categories. HERMES achieves the highest global
RFS and obtains the best performance among the selected comparison methods
in ten of the eleven categories. Particularly strong performance is observed in Single-Lane Maneuvers and
Special Vehicles, where HERMES achieves RFS scores of 8.1446 and 7.9264,
respectively. More broadly, the consistent gains across categories involving
heterogeneous road users, temporary road structures, unusual objects, and
complex maneuvering conditions indicate that the proposed multimodal and
risk-aware design is effective across diverse long-tail driving scenarios
rather than being specialized to a single scenario type.

\begin{table}[t]
\centering
\caption{Overall performance comparison on the official WOD-E2E test set}
\label{tab:overall_results}
\small
\begin{tabular}{lccc}
\toprule
\textbf{Method} &
\textbf{RFS}$\uparrow$ &
\textbf{ADE@3s}$\downarrow$ &
\textbf{ADE@5s}$\downarrow$
\tabularnewline
\midrule

OpenEMMA (\cite{xing2025openemma})
& 6.3162 & 1.8663 & 4.1078
\tabularnewline

LightEMMA~(\cite{qiao2025lightemma})
& 6.5169 & 1.7052 & 3.7395
\tabularnewline

AutoVLA~(\cite{zhou2025autovla})
& 6.9377 & 1.6207 & 3.5945
\tabularnewline


Dream-VLA (\cite{ye2025dreamvla})
& 7.1650 & 2.8687 & 4.1496
\tabularnewline


\textbf{HERMES (Ours)}
& \textbf{7.3915} & \textbf{1.5949} & \textbf{3.3060}
\tabularnewline
\bottomrule
\end{tabular}
\end{table}

\begin{table*}[t]
\centering
\caption{Category-wise RFS comparison on the official WOD-E2E test set. The eleven scenario categories include Spotlight, Construction, Intersection, Pedestrian, Cyclists, Multi-Lane Maneuvers, Single-Lane Maneuvers, Cut-ins, Foreign Object Debris (FOD), Special Vehicles, and Others.}
\label{tab:category_results}
\resizebox{\textwidth}{!}{
\begin{tabular}{lcccccccccccc}
\toprule
\textbf{Method} &
\textbf{Global} &
\textbf{Spotlight} &
\textbf{Construction} &
\textbf{Intersection} &
\textbf{Pedestrian} &
\textbf{Cyclists} &
\textbf{Multi-Lane} &
\textbf{Single-Lane} &
\textbf{Cut-ins} &
\textbf{FOD} &
\textbf{Special Vehicles} &
\textbf{Others}
\tabularnewline
\midrule


OpenEMMA (\cite{xing2025openemma})
& 6.3162 & 5.5054 & 6.3312 & 6.7374 & 6.6844 & 5.9696
& 6.1858 & 6.5395 & 6.4842 & 6.0835 & 6.7792 & 6.1786
\tabularnewline

LightEMMA~(\cite{qiao2025lightemma})
& 6.5169 & 5.7103 & 7.1768 & 6.5039 & 6.3576 & 6.5233
& 6.2180 & 7.2659 & 6.7744 & 7.0588 & 6.3521 & 5.7453
\tabularnewline

AutoVLA~(\cite{zhou2025autovla})
& 6.9377 & 5.6216 & 7.4599 & 7.1963 & 7.1731 & 6.9206
& 6.8144 & 7.3033 & 7.0362 & 7.0684 & 7.3669 & 6.3542
\tabularnewline


Dream-VLA (\cite{ye2025dreamvla})
& 7.1650 & 5.7591 & 7.2924 & 7.4669 & 7.5941 & \textbf{7.4017}
& 6.9261 & 7.9146 & 7.1132 & 7.0129 & 7.4448 & 6.8894
\tabularnewline


\textbf{HERMES (Ours)}
& \textbf{7.3915} & \textbf{6.4390} & \textbf{7.6561} & \textbf{7.4809}
& \textbf{7.6459} & 7.1708 & \textbf{7.0866} & \textbf{8.1446}
& \textbf{7.3701} & \textbf{7.3994} & \textbf{7.9264} & \textbf{6.9865}
\tabularnewline

\bottomrule
\end{tabular}
}
\end{table*}

\subsection{Ablation Study}
\label{sec:ablation}

As stated before, ablation studies are conducted on the WOD-E2E validation set to examine the
individual contributions of long-tail semantic instructions, high-level intent,
and historical state information in the proposed tri-modal driving framework.
All variants are trained and evaluated under the same experimental protocol as
the full model.

\subsubsection{No Instruction}

In this setting, we remove the entire \textit{Long-Tail Instruction Embedding}
module and its associated downstream pathway, including the
\textit{Scene Text Projector}, \textit{Scene Fusion},
\textit{Planning Text Projector}, and \textit{Risk Planning Cross-Attention}
modules. The planning representation is instead obtained by directly combining
visual features from the \textit{Vision Encoder} with temporal features from
the \textit{State Encoder}, followed by intent modulation and temporal decoding.

As shown in Table~\ref{tab:ablation}, removing the long-tail semantic
instructions reduces RFS from 6.81 to 6.21 and increases trajectory errors
across the reported ADE and FDE metrics. These results indicate that the
structured semantic guidance contributes to both human-preference-aligned
planning behavior and trajectory accuracy beyond visual and temporal cues alone.

\subsubsection{No Intent}

This variant removes the \textit{Intent Modulator} while retaining visual,
historical state, and semantic instruction inputs. The planning context produced
by \textit{Planning Context Fusion} is directly passed to the
\textit{Risk Planning Cross-Attention} module without explicit intent
conditioning.

Removing high-level intent decreases RFS from 6.81 to 6.56 and also increases
both ADE and FDE. Although the degradation is smaller than that observed when
semantic instructions are removed, the results show that explicit intent
conditioning contributes to maneuver-level planning and improves trajectory
consistency.

\subsubsection{No State}

In this setting, the \textit{State Encoder} is removed and historical ego-motion
information is no longer available. Consequently, \textit{Planning Context Fusion}
is omitted, and the scene-aware representation produced by \textit{Scene Fusion}
is directly passed to the \textit{Intent Modulator} and subsequent modules.

Removing historical state information produces the largest performance
degradation, with RFS decreasing from 6.81 to 5.51 and substantial increases
in both ADE and FDE. This result highlights the importance of temporal
ego-motion information for capturing vehicle dynamics and maintaining accurate
and stable trajectory prediction.

\subsubsection{Discussion}

Overall, the ablation results demonstrate complementary contributions from all
three information sources. Long-tail semantic instructions enhance risk-aware
and human-preference-aligned planning, high-level intent provides explicit
maneuver guidance, and historical state information supplies the temporal
motion context required for accurate trajectory prediction. The full HERMES
model achieves the best performance across all reported validation-set metrics,
demonstrating the benefit of jointly integrating these complementary signals.

\begin{table}[t]
\centering
\caption{Results of the ablation study on the WOD-E2E validation set}
\label{tab:ablation}
\small
\begin{tabular}{lccccc}
\toprule
\textbf{Method} &
\textbf{RFS}$\uparrow$ &
\textbf{ADE@3s}$\downarrow$ &
\textbf{ADE@5s}$\downarrow$ &
\textbf{FDE@3s}$\downarrow$ &
\textbf{FDE@5s}$\downarrow$
\tabularnewline
\midrule

No Instruction
& 6.21 & 0.92 & 1.94 & 1.98 & 4.82
\tabularnewline

No Intent
& 6.56 & 0.91 & 1.94 & 2.00 & 4.82
\tabularnewline

No State
& 5.51 & 2.48 & 4.22 & 4.69 & 8.59
\tabularnewline

\textbf{Base (HERMES)}
& \textbf{6.81}
& \textbf{0.82}
& \textbf{1.82}
& \textbf{1.73}
& \textbf{4.81}
\tabularnewline

\bottomrule
\end{tabular}
\end{table}

\begin{figure*}[]
  \centering
  \includegraphics[width=0.8\linewidth]{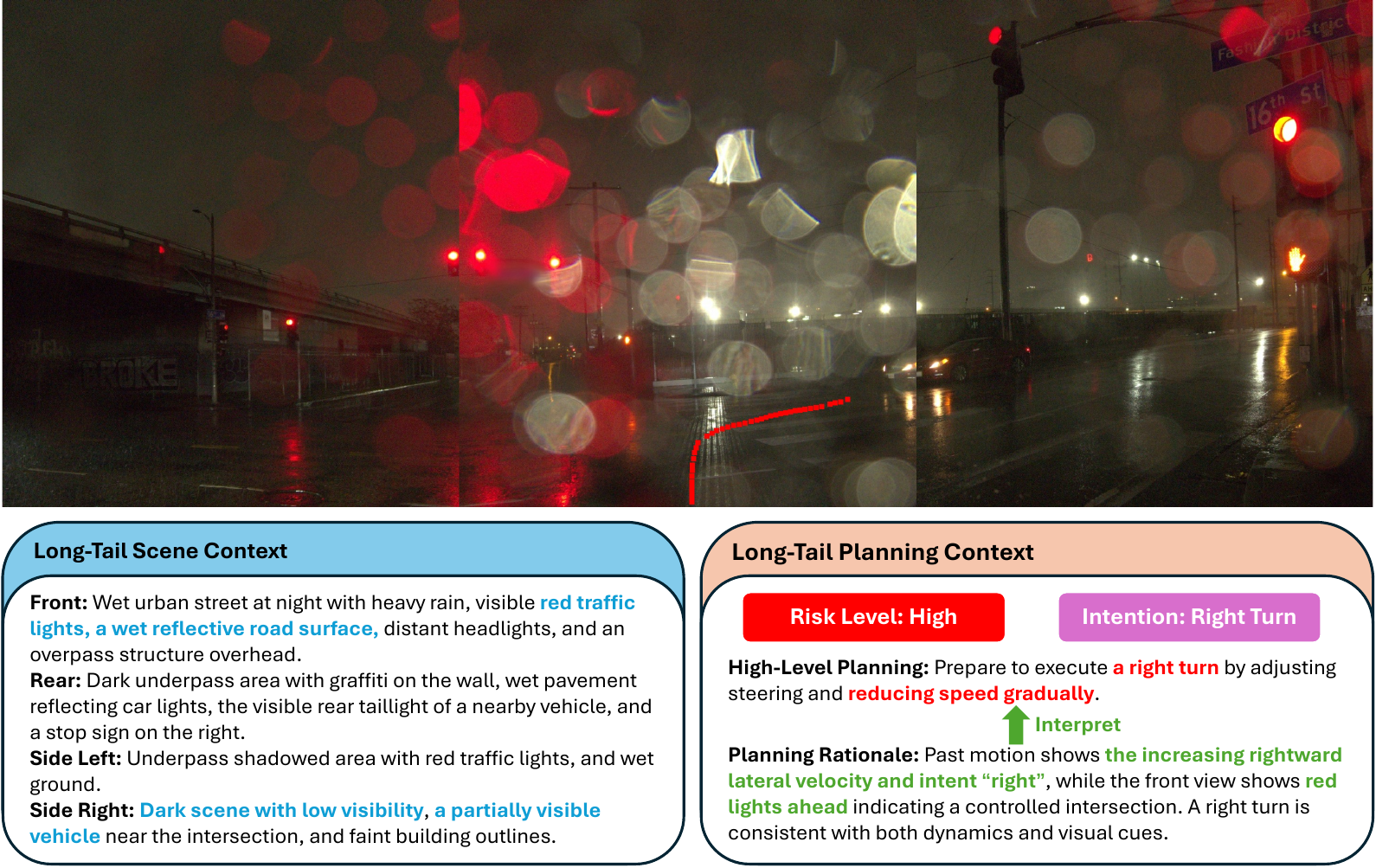}
  \caption{Nighttime driving under heavy rain and poor visibility. Planned trajectory is shown in red.}
  \label{fig:qual_case1}
\end{figure*}

\begin{figure*}[]
  \centering
  \includegraphics[width=0.8\linewidth]{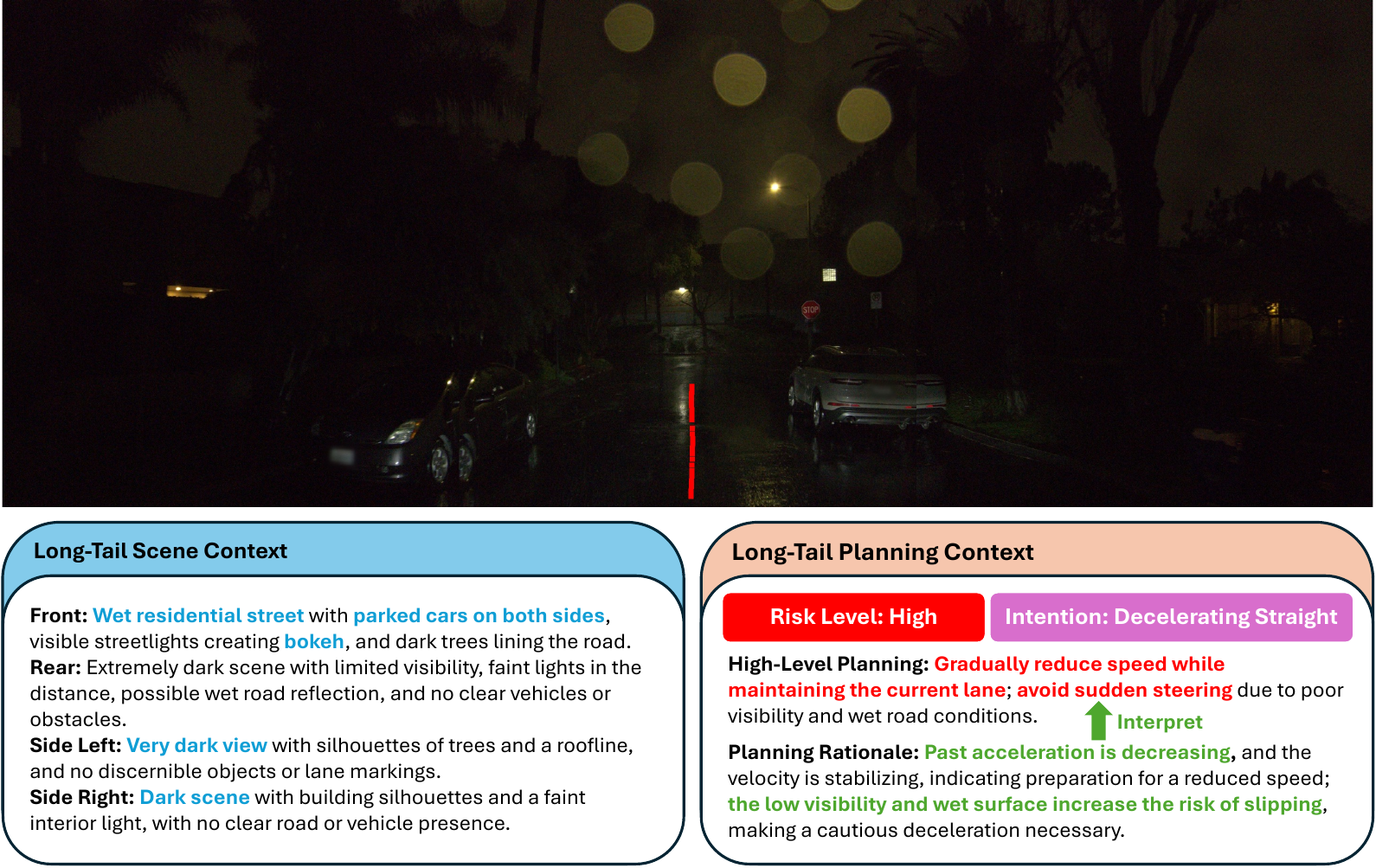}
  \caption{Extremely low-visibility residential street with wet road conditions. Planned trajectory is shown in red.}
  \label{fig:qual_case2}
\end{figure*}

\subsection{Qualitative Analysis}

In this section, we present qualitative examples to illustrate how HERMES leverages long-tail semantic instructions to produce robust and interpretable planning behaviors under challenging driving conditions.

\subsubsection{Case 1: Adverse Weather and Low Visibility}
This scenario depicts nighttime driving under heavy rain, where reflections on 
wet road surfaces and reduced illumination significantly degrade visual cues. 
Guided by the \textit{Long-Tail Scene Context} identifying low visibility and 
adverse weather conditions, HERMES generates a conservative and smooth turning 
trajectory with controlled speed, avoiding abrupt steering or acceleration. 
This example highlights the model’s ability to incorporate risk-aware semantic 
guidance when visual information alone is unreliable.

\subsubsection{Case 2: Residential Street under Extremely Degraded Visual Conditions}
In this example, the ego vehicle navigates a residential street with significantly limited 
visibility and wet pavement. Although no immediate obstacle is present, the 
scene requires cautious planning due to extremely dark scene, potential pedestrians and reduced 
friction. HERMES produces a stable and centered trajectory, reflecting a 
risk-aware driving strategy that balances progress and safety.

\subsubsection{Case 3: Construction Zone with Lane Channelization}
This scenario contains a work zone with barricades and temporary lane markings, 
resulting in an uncommon road layout. Such configurations are sparse in standard 
driving data and often challenge end-to-end planners. By leveraging the 
\textit{Long-Tail Planning Context}, HERMES anticipates the upcoming lane change 
and generates a smooth lateral maneuver while maintaining a controlled speed.

\subsubsection{Case 4: Complex Urban Intersection.}
This example illustrates a wide urban intersection with crosswalks and potential 
pedestrian interactions. Although the immediate path is unobstructed, the scene 
demands cautious behavior due to latent risks. HERMES generates a stable straight 
trajectory while maintaining readiness to yield, demonstrating appropriate 
planning behavior in moderately risky urban environments.

\begin{figure*}[]
  \centering
  \includegraphics[width=0.8\linewidth]{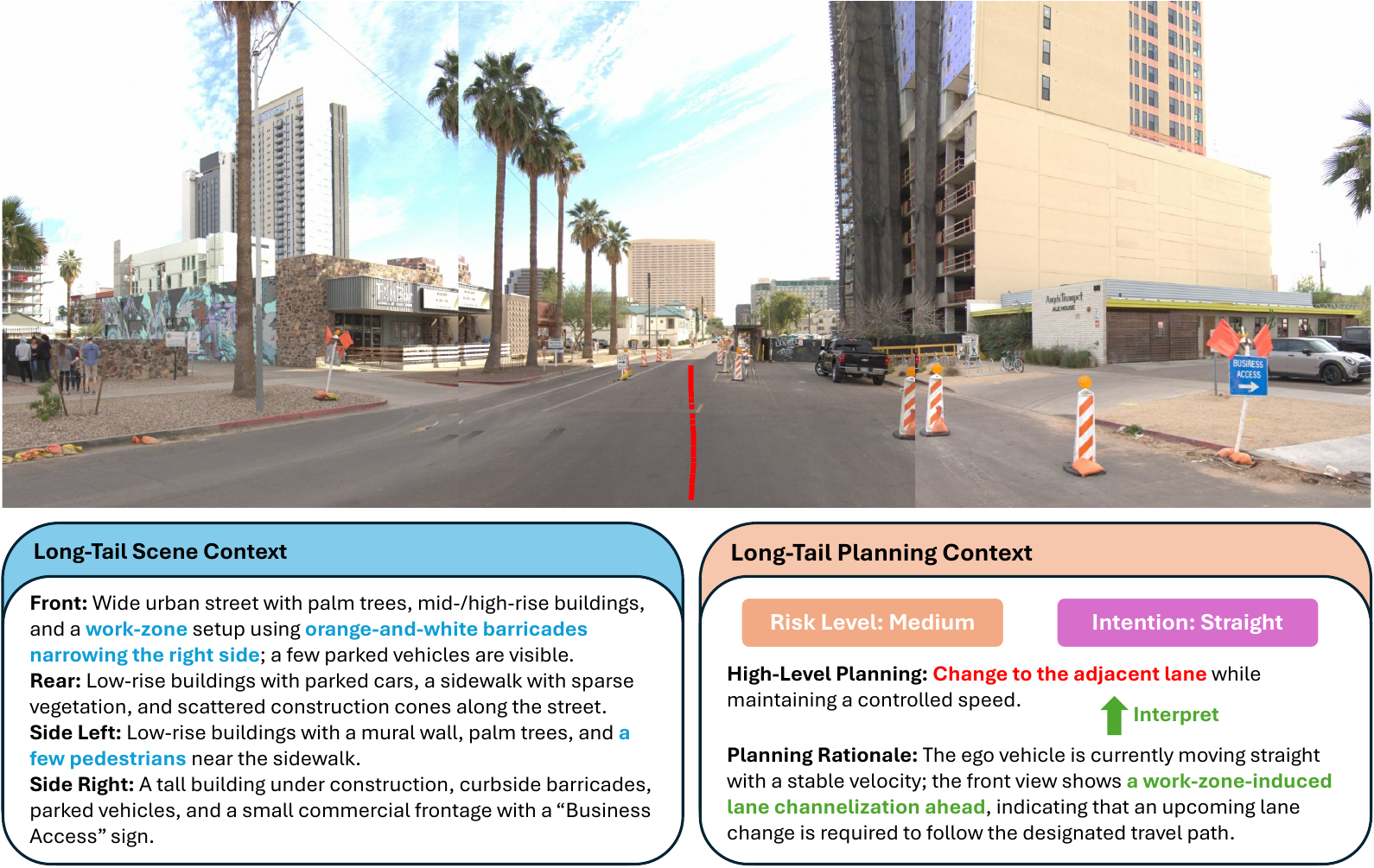}
  \caption{Work-zone scenario with lane channelization caused by construction. Planned trajectory is shown in red.}
  \label{fig:qual_case3}
\end{figure*}

\begin{figure*}[]
  \centering
  \includegraphics[width=0.8\linewidth]{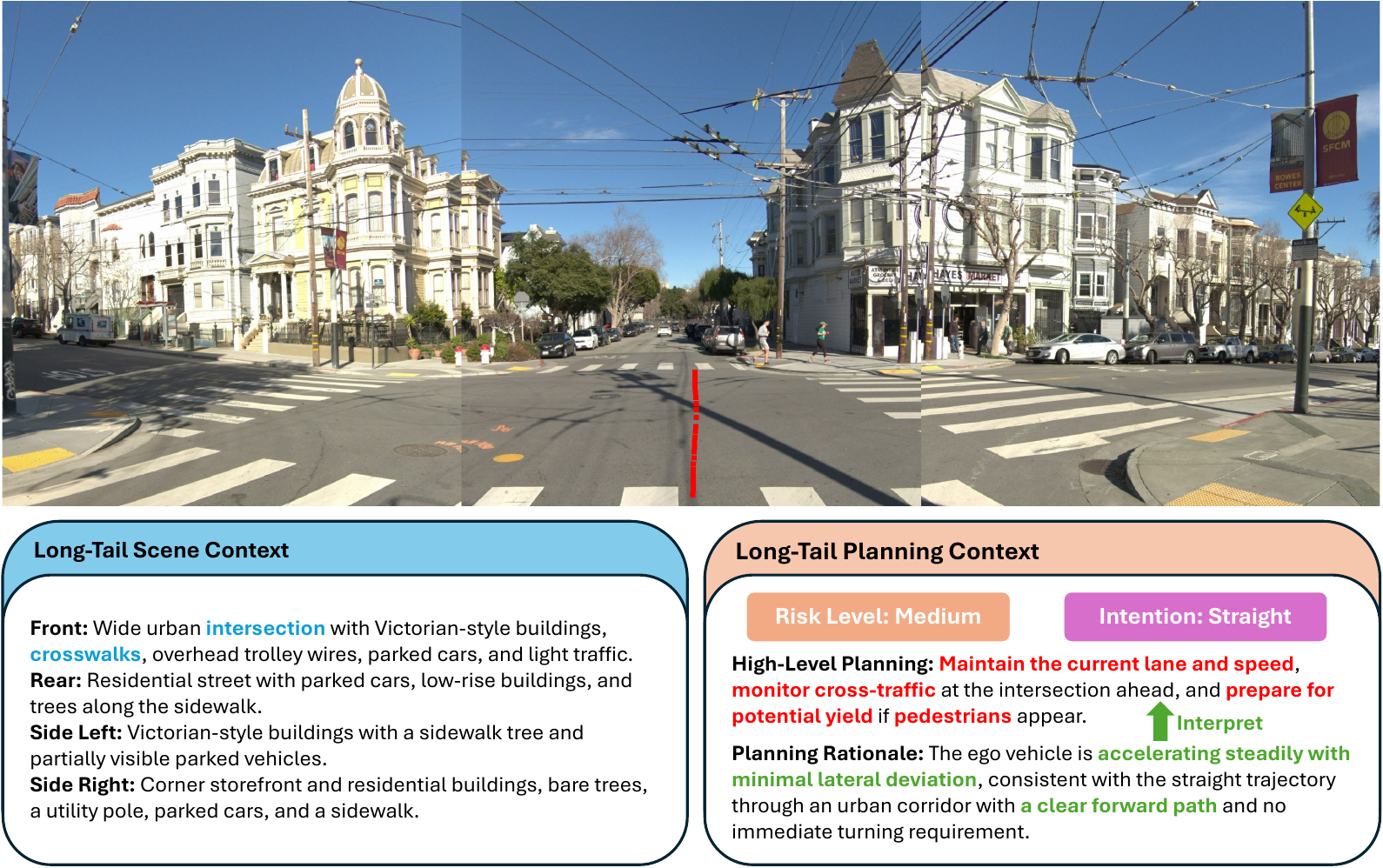}
  \caption{Complex urban intersection with crosswalks and surrounding traffic. Planned trajectory is shown in red.}
  \label{fig:qual_case4}
\end{figure*}

\section{Conclusion}

In this paper, we introduced HERMES, a holistic risk-aware end-to-end
multimodal autonomous driving framework designed to generate risk-aware and
accurate trajectories in safety-critical long-tail scenarios. HERMES
incorporates a foundation-model-assisted annotation pipeline that produces
structured scene and planning contexts and explicitly integrates these signals
into end-to-end trajectory generation. Building on this semantic guidance,
the proposed tri-modal planning architecture jointly fuses multi-view
perception, historical motion cues, and long-tail context through risk- and
intent-aware conditioning, supporting risk-aware planning while maintaining
trajectory accuracy.

Experiments show that HERMES achieves the
best overall performance among the state-of-the-art methods. It also
maintains the highest RFS in ten critical scenarios,
indicating consistent performance across diverse long-tail driving conditions
involving heterogeneous road users, uncommon hazards, and complex driving
interactions. The ablation study further shows that removing
long-tail semantic instructions, high-level intent, or historical state
information leads to performance degradation. Together, these results
demonstrate that structured long-tail semantic guidance serves as an effective
planning signal rather than merely auxiliary annotation, and validate the
effectiveness of HERMES for risk-aware end-to-end driving.

Future work will focus on improving annotation reliability under ambiguity
and extending evaluation to closed-loop settings with explicit safety
constraints. Moreover, integrating long-tail scenario generation and
world-model-based high-fidelity simulation can further strengthen robustness to complex and
rapidly changing driving conditions.

\printcredits

\bibliographystyle{cas-model2-names}

\bibliography{citation}

\end{document}